# Testing report of a fingerprint-based door-opening system[1]

Marcos Faundez-Zanuy, Joan Fabregas
Escola Universitaria Politècnica de Mataró
Avda. Puig i Cadafalch 101-111
08303 MATARO (BARCELONA) SPAIN
E-mail: faundez@eupmt.es, fabregas@eupmt.es http:www.eupmt.es/veu

**ABSTRACT**
This paper describes the operational evaluation of a door-opening system based on a low-cost inkless fingerprint sensor. This system has been developed and installed for access control to one of our laboratories. Experimental results reveal that the system is working fine and no special cleaning requirements neither components replacement is needed. It can support more than 50 users, and an average of 74,5 access attempts per day in a 14-hour 5-day-per-week working.
Emphasize is also given on some important facts to be taken into consideration when comparing and evaluating different products from different vendors.

## TESTING A BIOMETRIC DEVICE

When testing a biometric device, special care must be taken about the conditions in which the results have been obtained. Let us think, for instance, on a comparison of three hypothetical products from three different vendors. We can use several parameters that model the behavior of a verification system: if a verification system accepts an impostor, it makes a False Acceptance (FA) error. If the system rejects a valid user, it makes a False Rejection (FR) error. Both errors can be traded off by adjusting the decision threshold. The operating point where the FA and FR are equal corresponds to the Equal Error Rate (EER). Table 1, first column, reports the Equal Error Rate (EER) obtained from the datasheets.

|  | EER | | |
|---|---|---|---|
|  | datasheet | Elder people | After half a year |
| Product 1 | 0.5% | 0.5% | 2% |
| Product 2 | 0.05% | 3% | 2% |
| Product 3 | 1% | 1% | 1% |

**Table 1. Comparison between three hypothetical systems for fingerprint access control.**

Which is the best device? According to this information, doubtless, the second one. However, we decide to install the system in a old people's home, and for this target population we get the taxes shown on second column. We repeat the experiment half a year later, and we get the last column results. Now, with perspective, which is the best system?. Probably, taking into account the whole information of table 1, the third product. Unfortunately, we do not know how the results have been computed by the manufacturer, but probably they have been obtained in optimal conditions. In addition, these conditions vary along the experiments of different makers, so it is hard (if not impossible) to compare different products. There are hundreds of popular sentences related to statistics [2], which can be considered a double-edge knife. Some examples are:
- Gregg Easterbrook: "Torture numbers, and they will confess to anything".
- Andrew Lang: "He uses statistics as a drunken man uses lamp posts - for support rather than for illumination".

While there are some regulations, for instance, for car fuel consumption and Olympic Games (it is not the same tail/head wind, steep up/down, etc.), there is neither regulation nor standardization for testing biometric devices, where little information about the experimental conditions is given.
For this reason, it is especially interesting to get information from impartial entities without economical interests on a given product.

---

[1] This work has been supported by FEDER and MCYT, TIC-2003-08382-C05-02

**OPERATIONAL EVALUATION OF THE DOOR OPENING SYSTEM**
Although there are neither standard rules nor an official homologation laboratory, some best practices in testing and reporting performance of biometric devices exist [3]. According to them, three different testing levels can be established:

**Technology evaluation:** The goal of a technology evaluation is to compare competing algorithms from a single technology. Testing of all algorithms is carried out on a standardized database collected by a "universal" sensor. Nonetheless, performance with this database will depend upon both the environment and the population in which it is collected. Testing is carried out using offline processing of the data. Because the database is fixed, the results of technology tests are repeatable.

**Scenario evaluation:** The goal of scenario testing is to determine the overall system performance in a prototype or simulated application. Testing is carried out on a complete system in an environment that models a real-world target application of interest. Each tested system will have its own acquisition sensor and so will receive slightly different data. Consequently, care will be required that data collection across all tested systems is in the same environment with the same population. Test results will be repeatable only to the extent that the modeled scenario can be carefully controlled.

**Operational evaluation:** The goal of operational testing is to determine the performance of a complete biometric system in a specific application environment with a specific target population. Depending upon data storage capabilities of the tested device, offline testing might not be possible. In general, operational test results will not be repeatable because of unknown and undocumented differences between operational environments. Further, "ground truth" (i.e. who was actually presenting a "good faith" biometric measure) will be difficult to ascertain.

According to this classification, our experiments correspond to the latter class. In operational testing, the environment and the population are determined in situ with little control over them by the experimenter. In our case, the target population is students (ranging from 18 to 22 years old) plus lecturers (ranging from 35 to 60 years old).
In our setting up [1] the user does not need to provide his/ her claimed identity, because the system works on identification mode. The input fingerprint is matched against the whole database, and two main kinds of errors can result:
> **False Acceptance (FA):** when a submitted fingerprint belonging to a person outside the database is incorrectly matched to a template enrolled by another user. In this situation the door is incorrectly opened.
> **False Rejection (FR):** when a submitted fingerprint belonging to a person in the database is neither close enough to a template enrolled previously by him/ her nor any other user (these two cases cannot be differentiated, although the former will be clearly more frequent). In this situation the door is not opened.
> Taking into account that there is no human supervisor of the system, False Acceptances cannot be identified.

We track information about time of attempt to access and the name of identified person or unknown if it is not found on the database.
Without a human supervisor, some criterion must be taken in order to detect the different situations.
Basically, we can consider that the system collects this information:
> **Failure to enroll:** Measures the proportion of volunteers who could not be enrolled. Taking into account that there is human supervision for the enrolment of new users, this amount can be experimentally determined. In our system, we have not found any person unable to enroll.

**Failure to acquire:** is defined as the expected proportion of transactions for which the system is unable to capture or locate an image of sufficient quality. Fortunately, the fingerprint sensor U.are.U [4] does not take any fingerprint unless it has enough quality.

**Acceptances:** This amount is automatically detected from the number of times that the door is opened. Unfortunately, there is no human being supervising the system, so it is not possible to differentiate the number of False Acceptances from the number of Genuine Acceptances. Thus, it includes both amounts.

**Rejections:** This amount is automatically detected from the number of times that the door is not opened. Unfortunately, there is no human being supervising the system, so it is not possible to differentiate the number of false rejections from the number of fake users attempting to enter the system. However, we will consider that a false rejected person will try to enter again, and will repeat the trial. Thus, a rejection followed by a success attempt in less than 2 seconds will be considered as a false rejection.

The number of impostors trying to enter the facilities is expected to be higher at the beginning of the system installation, and smaller as the time goes.

## RESULTS

In [1] we presented a design for a door-opening system using a low-cost fingerprint scanner and a personal computer. This system lets us control the students' access to the laboratory, so they do not need any more to have a key nor to wait for someone to open the door. It lets rationalize accesses, update the authorized persons inside the database without collecting or distributing any physical keys, etc.

The door-opening system has also allowed us to check that, having a proper configuration, it is able to work during the whole laboring time, and it also permits studying the number of accesses and users that it is supporting. Up to now the database consists of 61 persons and near 10.000 accesses. Based on the data collection of these users, the following results (table 2) have been achieved:

| Parameter | Value |
|---|---|
| Number of accesses | 6896 |
| Failure to Enroll Rate | 0 % |
| Acceptances | 52,39 % |
| Rejections | 47,61 % |
| False Rejections | 0,4 % |
| Average accesses per day | 74,49 |
| Minimum accesses per day | 4 |
| Maximum accesses per day | 195 |

**Table 2. Experimental results. Rejections includes the number of impostors trying to grant access, while False Rejection considers only genuine users.**

In addition, we have checked that, after more than 6 months of continuous operation mode, special cleaning or component replacement had not to be done. Thus, we consider that this kind of device is suitable for this kind of applications, and we encourage potential users with similar requirements.